\documentclass{article}

\usepackage{microtype}
\usepackage{graphicx}
\usepackage{subcaption}
\usepackage{booktabs} 

\usepackage{hyperref}

\usepackage[preprint]{icml2026}

\usepackage{amsmath}
\usepackage{amssymb}
\usepackage{mathtools}
\usepackage{amsthm}

\usepackage[capitalize,noabbrev]{cleveref}

\theoremstyle{plain}

\theoremstyle{definition}

\theoremstyle{remark}

\usepackage[textsize=tiny]{todonotes}

\icmltitlerunning{Self-Assessment Reinforcement Learning for Search-State Awareness in Retrieval-Augmented Large Language Models}

\usepackage{tabularx}
\usepackage{booktabs}
\usepackage{multirow}
\usepackage{colortbl}
\usepackage{xcolor}
\usepackage{listings}

\usepackage{graphicx}
\usepackage{subcaption}
\usepackage{booktabs} 

\usepackage{amsmath}
\usepackage{amssymb}
\usepackage{mathtools}
\usepackage{amsthm}
\usepackage{float}
\begin{document}

\twocolumn[
  \icmltitle{Self-Assessment Reinforcement Learning for Search-State Awareness in Retrieval-Augmented Large Language Models}



  \icmlsetsymbol{prior}{*}

  \begin{icmlauthorlist}
    \icmlauthor{Taewhoo Lee}{sch,comp,prior}
    \icmlauthor{Minju Song}{sch}
    \icmlauthor{Jungwoo Park}{sch,comp2}
    \icmlauthor{Chanwoong Yoon}{sch2}
    \icmlauthor{Junha Jung}{sch,comp2}
    \icmlauthor{Hyeon Hwang}{sch}
    \icmlauthor{Sanghoon Lee}{sch,comp2}
    \icmlauthor{Minbyul Jeong}{comp3}
    \icmlauthor{Jaewoo Kang}{sch,comp2}
  \end{icmlauthorlist}

  \icmlaffiliation{comp}{Amazon}
  \icmlaffiliation{comp2}{AIGEN Sciences}
  \icmlaffiliation{comp3}{Meridian Intelligence Global Inc}
  \icmlaffiliation{sch}{Korea University}
  \icmlaffiliation{sch2}{University of Massachusetts Amherst}

  \icmlcorrespondingauthor{Minbyul Jeong}{wjdalsquf@gmail.com}
  \icmlcorrespondingauthor{Jaewoo Kang}{kangj@korea.ac.kr}

  \icmlkeywords{Machine Learning, ICML}

  \vskip 0.3in
]



\printAffiliationsAndNotice{\textsuperscript{*}Work done prior to joining Amazon.\newline}  

\begin{abstract}
Recent advances in search agents have enabled large language models to tackle complex tasks by retrieving and synthesizing external information. However, existing search agents often lack search-state awareness, i.e., the ability to recognize what information has been obtained and what remains missing, leading to hallucinated reasoning and prematurely terminated search.
To address this limitation, we propose Self-Assessment Reinforcement Learning (SARL), a reinforcement learning framework in which models learn to recognize their own search state during training and improve their search behavior accordingly, without requiring additional assessment at test time.
SARL achieves this through a dual reward design that captures the corrective effect of each assessment while penalizing excessive reliance on it, jointly encouraging effective and efficient search trajectories.
Under this setup, SARL achieves a 24.3\% relative improvement over the strongest baseline, with average gains of 35.9\% on multi-hop reasoning and 27.9\% on deep research benchmarks against the strongest per-benchmark baselines.
SARL further surpasses methods built on auxiliary supervision or multi-agent coordination, despite operating as a standalone RL framework with no inference-time scaffolding.
Beyond final performance, SARL maintains training stability under extended search horizons, a regime where existing methods consistently degrade, enabling sustained gains as task complexity grows.
\end{abstract}

\section{Introduction}
\label{section:introduction}

Large language models (LLMs) have shown remarkable performance across a wide range of tasks, but their reliance on static parametric knowledge limits their ability to handle up-to-date information~\citep{lewis2020retrieval, guu2020retrieval}.
Retrieval-augmented generation (RAG) alleviates this by equipping LLMs with external information at inference time~\citep{asai-etal-2023-retrieval, snell2024scaling, gao2023retrieval}, and recent \emph{search agents} extend this capability by enabling models to autonomously interact with external environments, iteratively performing searches and synthesizing evidence from multiple sources~\citep{jin2025search, Search-o1}.
However, naturally incorporating this behavior into LLMs remains challenging due to the large action space involved during search~\citep{yao2022react, nakano2021webgpt, schick2023toolformer}. While supervised fine-tuning (SFT) using high-quality search trajectories has proven effective~\citep{he2025webseer, wu2025webdancer, li2025websailor, Li2025WebThinker, liu2025webexplorer}, it requires resource overhead from collecting intermediate search decisions and faces difficulties in generalizability~\citep{chu2025sft}. Reinforcement learning (RL) has emerged as a strategy to teach agents effective search strategies through interaction and outcome-based feedback, reducing reliance on expensive human annotations~\citep{jin2025search, shi2025search, song2025r1, song2025r1++}.

\begin{figure*}[t]
\centering
\includegraphics[width=\textwidth]{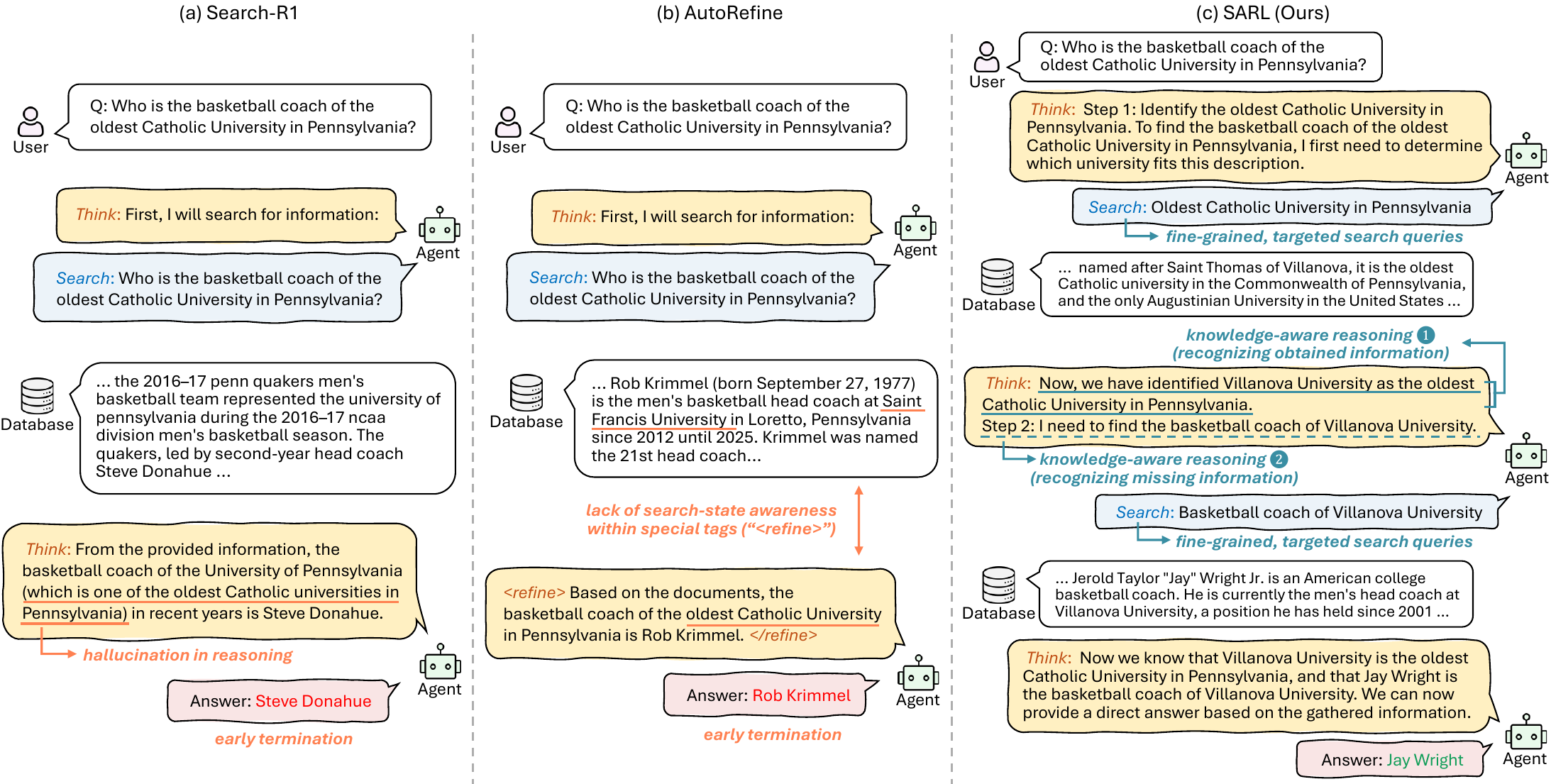}
\caption{
Comparison of search trajectories among Search-R1~\citep{jin2025search}, AutoRefine~\citep{shi2025search}, and SARL (Ours). 
(a) Search-R1 hallucinates by assuming the University of Pennsylvania is a Catholic institution and terminates early. 
(b) AutoRefine fails to recognize that Saint Francis University does not satisfy the ``oldest Catholic University'' constraint, despite using a dedicated refinement tag. 
(c) SARL identifies the correct entity (Villanova University) by recognizing what has been obtained and what remains missing, issuing targeted queries to fill the gap.
}
\label{figure:motivation_figure}
\end{figure*}

Despite these advances, current RL-based search agents struggle with \emph{search-state awareness}: the ability to recognize what information has been obtained and what remains missing during complex retrieval. Without this awareness, agents fall into two failure modes: they \emph{hallucinate reasoning}, generating answers from incomplete evidence (e.g., asserting that the University of Pennsylvania is ``a Catholic institution'' in Figure~\ref{figure:motivation_figure}), and they \emph{prematurely terminate}, stopping searches before sufficient evidence is gathered.%
\footnote{We provide failure mode illustrations in Appendix~\ref{appendix:trajectory_examples}.}
Existing methods attempt to mitigate these issues through sparse outcome-based rewards or auxiliary tokens (e.g., \texttt{<refine>}, \texttt{<judge>}), but such training signals are coarse and fail to ensure faithfulness of the reasoning process~\citep{xu2025beyond, zhang2025criticsearch}. We further find that these methods exhibit severe training instability and performance degradation under extended search horizons, a failure mode unreported in prior work.

\begin{figure*}[t]
\centering
\includegraphics[width=0.95\textwidth]{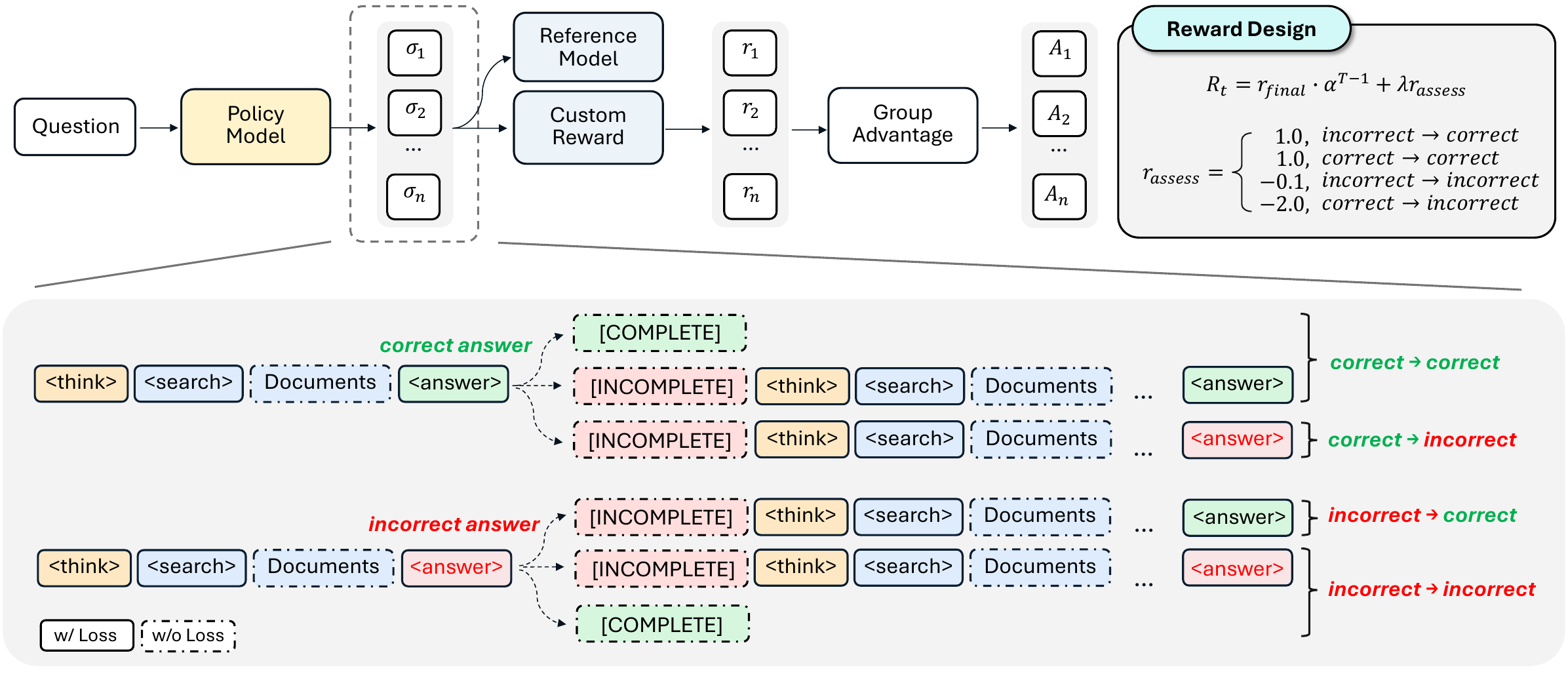}
\caption{
Overview of the SARL framework. 
\textbf{Top:} A group of trajectories sampled from the policy model is scored under our custom reward (Eq.~\ref{eq:reward}) and converted to group-relative advantages via GRPO~\citep{shao2024deepseekmath}.
\textbf{Bottom:} Each trajectory ends with a structured self-assessment, which resolves to a \texttt{COMPLETE} or \texttt{INCOMPLETE} judgment; \texttt{INCOMPLETE} triggers trajectory revision, and $r_{\mathrm{assess}}$ is determined by the resulting transition in answer correctness. Solid borders indicate trajectory components included in the loss; dashed borders are excluded.
}
\label{figure:main_figure}
\end{figure*}

To overcome these limitations, we propose \textbf{Self-Assessment Reinforcement Learning (SARL)}, an RL framework that cultivates search agents with stronger search-state awareness. SARL is grounded in the premise that models can identify and correct their own erroneous trajectories when given a fresh context to review them (validated in Appendix~\ref{appendix:central_premise}). 
Building on this premise, SARL extends the standard think-search-answer loop with a \emph{self-assessment phase}: after an initial answer, the model re-examines its own trajectory in an isolated context and produces a structured assessment that identifies the essential constraints of the question, analyzes which have been resolved by the retrieved evidence, and concludes with a judgment on whether the trajectory is complete. When the assessment indicates incompleteness, the full assessment is injected back as user feedback to prompt revision; otherwise the trajectory terminates.

To shape this behavior into an internalized capability, SARL introduces a dual reward mechanism. An \emph{assessment reward} captures the corrective effect of each assessment, computed from the transition in answer correctness before and after the assessment step. An \emph{exponential assessment penalty} bounds assessment frequency, encouraging thorough searching upfront rather than repeated reliance on assessment retries. Together, these signals push the model to correct mistakes when necessary while progressively internalizing search-state awareness, leading to accurate search behavior without explicit assessment steps at inference time.

We evaluate SARL across ten benchmarks spanning three complexity levels: three single-hop question answering (QA), four multi-hop QA, and three deep research benchmarks. To ensure fair comparison, we train all methods on a unified 45K-sample training set under a fixed maximum of 32 turns. 
SARL achieves a 24.3\% relative improvement over the strongest baseline, with average gains of 35.9\% on multi-hop reasoning and 27.9\% on deep research benchmarks against the strongest per-benchmark baselines.
SARL further surpasses methods from broader paradigms, including SFT+RL pipelines and multi-agent frameworks with larger backbones, despite operating as a standalone RL framework with no inference-time scaffolding. Our analysis further reveals that SARL progressively learns to identify and fill information gaps through diverse search queries with notably higher evidence recall.

We summarize our main contributions as follows:
\begin{itemize}
\item We identify \emph{search-state awareness} as a core capability missing in current RL-based search agents, and surface a previously unreported failure mode of training instability under extended search horizons.
\item We propose \textbf{SARL}, an RL framework that internalizes search-state awareness through a train-time self-assessment phase shaped by a dual reward mechanism, requiring no auxiliary supervision, multi-agent orchestration, or inference-time overhead.
\item Across ten benchmarks spanning three complexity levels, SARL achieves a 24.3\% average improvement over the strongest baseline with stable training and strong generalizability across model scales and families.
\end{itemize}
\section{Method}
\label{section:Method}

\subsection{Problem Setup and Framework}
\label{subsec:task_definition}

\paragraph{Search-augmented question answering.}
We study the task of search-augmented question answering, where a language model is equipped with access to an external search engine $\mathcal{E}$. Given a dataset $\mathcal{D}=\{(q,a)\}$ of question-answer pairs, the model answers each question $q$ by generating an interaction trajectory $\mathcal{T}=\{(u_1,x_1),\dots,(u_K,x_K)\}$ of length $K$, where each step consists of an operation type $u_k\in\{\langle\text{think}\rangle,\langle\text{search}\rangle,\langle\text{answer}\rangle\}$ and its associated generated content $x_k$.
When the model emits a $\langle\text{search}\rangle$ operation, the enclosed query is executed by $\mathcal{E}$, which returns top-$k$ relevant documents that are appended to the context for subsequent steps. The trajectory terminates with an $\langle\text{answer}\rangle$ operation or when a predefined maximum number of turns $K_{\max}$ is reached, after which the generated answer is evaluated against the ground-truth answer $a$.

\paragraph{Self-assessment phase.}
SARL augments the standard generation procedure with a \emph{self-assessment phase} that operates on completed trajectories. Given a trajectory $\mathcal{T}$ ending in $\langle\text{answer}\rangle$, the same policy $\pi_\theta$ is invoked in an isolated context with an evaluation prompt $p_{\text{eval}}$ (Appendix~\ref{appendix:assessment_prompt}) to produce a structured assessment $\hat{s} = \pi_\theta(\mathcal{T},\, p_{\text{eval}})$, from which a binary judgment $s \in \{\texttt{COMPLETE}, \texttt{INCOMPLETE}\}$ is extracted.
By decoupling self-assessment from the search loop, the original trajectory $\mathcal{T}$ remains a self-contained record of the model's search behavior, with the assessment treated as a separate evaluative pass.

\paragraph{Trajectory revision.}
If $s = \texttt{COMPLETE}$, the trajectory is finalized and the answer is evaluated against $a$. If $s = \texttt{INCOMPLETE}$, $\hat{s}$ is injected as a user message back into the original context (Appendix~\ref{appendix:revision_prompt}), prompting $\pi_\theta$ to extend $\mathcal{T}$ with further $\langle\text{think}\rangle$ and $\langle\text{search}\rangle$ steps before producing a revised $\langle\text{answer}\rangle$, yielding $\mathcal{T}'$. The procedure repeats until either $s = \texttt{COMPLETE}$ or the trajectory reaches the maximum number of turns $K_{\max}$. Figure~\ref{figure:main_figure} provides an overview of the full framework.

\subsection{Dataset Curation}
\label{subsec:dataset_curation}

We curate a diverse training dataset covering search tasks of increasing complexity. For single-hop question answering, we randomly select 15K samples from Natural Questions (NQ)~\citep{kwiatkowski2019natural}, where most questions can be resolved with a single retrieval step. For multi-hop reasoning, we randomly select 15K samples from HotpotQA~\citep{yang2018hotpotqa}, which requires aggregating evidence across multiple documents. Finally, for more challenging deep-research scenarios, we include 15K samples from ASearcher~\citep{gao2025beyond}, a dataset designed for complex, multi-step information-seeking behavior. This split consists of 8{,}537 examples from the original \texttt{compose\_chain\_qa} subset provided by ASearcher, along with 6{,}463 additional examples that we generate using the data synthesis procedure from ASearcher. This construction yields a balanced corpus of 45K instances spanning single-hop retrieval, multi-hop reasoning, and long-horizon search trajectories.

\subsection{Training Objective}
\label{subsec:training_objective}

We adopt Group Relative Policy Optimization (GRPO)~\citep{shao2024deepseekmath} as our policy optimization algorithm. Under GRPO, for each question $q$, a group of $G$ trajectories $\{o_i\}_{i=1}^G$ is sampled from the current policy $\pi_\theta$, and policy updates are performed based on group-relative comparisons across trajectories. In addition, following~\citet{yu2025dapo}, we mask out gradient computation for trajectories exceeding the model's original context length, and adopt asymmetric clipping. Lastly, we remove standard deviation normalization, replace per-trajectory length normalization with a fixed denominator $L_{\max}$, and omit KL regularization~\citep{liu2025understanding}. The resulting objective is
\begin{equation}
\mathbb{E}\!\left[
\frac{1}{G}
\sum_{i=1}^G
m_i
\frac{1}{L_{\max}}
\sum_{t}
\min\!\left(
\rho_{i,t}A_{i,t},
\mathrm{clip}^{\pm}(\rho_{i,t})A_{i,t}
\right)
\right],
\end{equation}
where $\rho_{i,t}=\pi_\theta(o_{i,t})/\pi_{\theta_{\mathrm{old}}}(o_{i,t})$, $m_i$ is the trajectory-level loss mask, and $\mathrm{clip}^{\pm}(\rho)=\mathrm{clip}(\rho,1-\epsilon^{-},1+\epsilon^{+})$ with $\epsilon^{-}=0.2$ and $\epsilon^{+}=0.28$.
In SARL, $A_{i,t}$ is computed from the SARL-specific reward $R_{\text{total}}$ (Section~\ref{subsec:reward_design}), which combines the assessment penalty and assessment reward.

\subsection{Reward Design}
\label{subsec:reward_design}

When training search agents with self-assessment, we aim to satisfy two complementary objectives: (1) \textit{efficiency}, ensuring that the learned agent does not rely on self-assessments at inference time, and (2) \textit{effectiveness}, enabling the model to leverage self-assessments during training to learn complete trajectories with search-state awareness. To jointly capture these objectives, we combine an assessment penalty with a fine-grained assessment reward.

For a trajectory containing $n$ self-assessment steps, the total reward is defined as
\begin{equation}
\label{eq:reward}
R_{\mathrm{total}} = r_{\mathrm{final}} \cdot \alpha^{n-1} + \lambda \, r_{\mathrm{assess}},
\end{equation}
where $r_{\mathrm{final}}$ denotes the final outcome reward, $\alpha \in (0,1)$ controls the \emph{assessment penalty}, and $r_{\mathrm{assess}}$ denotes the \emph{assessment reward}, weighted by coefficient $\lambda$. The exponential decay term $\alpha^{n-1}$ penalizes frequent self-assessments, encouraging the model to use assessments sparingly and to produce complete responses without relying on repeated retries, thereby discouraging dependence on self-assessment at inference time while preserving its utility during training.

To promote effective use of self-assessments, the assessment reward $r_{\mathrm{assess}}$ provides fine-grained feedback based on how the model's answer quality changes after an assessment. Specifically, it is determined by the transition in answer correctness before and after a self-assessment step:
\begin{equation}
\label{eq:feedback_reward}
r_{\mathrm{assess}} =
\begin{cases}
\phantom{-}1.0, & \text{incorrect} \rightarrow \text{correct}, \\
\phantom{-}1.0, & \text{correct} \rightarrow \text{correct}, \\
-0.1, & \text{incorrect} \rightarrow \text{incorrect}, \\
-2.0, & \text{correct} \rightarrow \text{incorrect}.
\end{cases}
\end{equation}
This asymmetric design strongly rewards corrective behavior, mildly penalizes unproductive self-assessments, and heavily discourages regressions in correctness. We assign identical rewards to \text{incorrect}$\rightarrow$\text{correct} and \text{correct}$\rightarrow$\text{correct} cases to avoid incentivizing the model to produce incorrect initial answers in pursuit of larger correction rewards. Moreover, the \text{correct}$\rightarrow$\text{correct} case often corresponds to situations where the answer remains correct but the supporting evidence is incomplete; rewarding this transition encourages the model to pursue well-supported search trajectories.
\section{Experiments}
\label{section:experiments}

\begin{table*}[t]
\centering
\scriptsize
\setlength{\tabcolsep}{3pt}
\renewcommand{\arraystretch}{1.05}
\caption{Performance comparison across benchmarks of varying complexity. NQ, 2Wiki, and BC-Plus denote Natural Questions, 2WikiMultihopQA, and BrowseComp-Plus, respectively. \textbf{Bold} indicates the best result per benchmark; \underline{underline} indicates the second-best.}
\begin{tabularx}{\textwidth}{
l
*{10}{>{\centering\arraybackslash}X}
c
}
\toprule
\multirow{2}{*}{\textbf{Methods}} 
& \multicolumn{3}{c}{\textbf{Single-Hop QA}} 
& \multicolumn{4}{c}{\textbf{Multi-Hop QA}} 
& \multicolumn{3}{c}{\textbf{Deep Research}} 
& \multirow{2}{*}{\textbf{Avg.}} \\
\cmidrule(lr){2-4} 
\cmidrule(lr){5-8}
\cmidrule(lr){9-11}
& \textbf{NQ} & \textbf{TriviaQA} & \textbf{PopQA}
& \textbf{HotpotQA} & \textbf{2Wiki} & \textbf{Musique} & \textbf{Bamboogle}
& \textbf{GAIA} & \textbf{FRAMES} & \textbf{BC-Plus}
&  \\
\midrule

\multicolumn{12}{l}{\textit{w/o Retrieval}} \\
\hspace{0.6em} Direct Generation
& 0.251 & 0.523 & 0.158 & 0.265 & 0.245 & 0.080 & 0.208 & 0.039 & 0.090 & 0.037 & 0.190 \\
\hspace{0.6em} SFT
& 0.181 & 0.452 & 0.161 & 0.217 & 0.212 & 0.048 & 0.126 & \textbf{0.053} & 0.062 & 0.003 & 0.152 \\
\hspace{0.6em} R1
& 0.233 & 0.543 & 0.200 & 0.271 & 0.305 & 0.084 & 0.232 & 0.046 & 0.121 & 0.006 & 0.148 \\
\midrule

\multicolumn{12}{l}{\textit{w/ Retrieval}} \\
\hspace{0.6em} Naive RAG
& 0.433 & 0.114 & 0.434 & 0.136 & 0.175 & 0.022 & 0.058 & 0.027 & 0.048 & 0.028 & 0.147 \\
\hspace{0.6em} IRCoT
& 0.381 & 0.127 & 0.417 & 0.148 & 0.201 & 0.019 & 0.068 & \underline{0.051} & 0.052 & 0.013 & 0.148 \\
\midrule

\multicolumn{12}{l}{\textit{Search Agents}} \\
\hspace{0.6em} Search-R1
& 0.459 & 0.711 & \underline{0.495} & 0.484 & \underline{0.440} & \underline{0.172} & 0.388 & 0.036 & 0.166 & 0.068 & 0.342 \\
\hspace{0.6em} ASearcher
& \textbf{0.515} & \underline{0.719} & 0.492 & \underline{0.520} & 0.390 & 0.161 & 0.344 & 0.039 & \underline{0.176} & \underline{0.097} & \underline{0.345} \\
\hspace{0.6em} AutoRefine
& 0.493 & 0.661 & 0.450 & 0.356 & 0.187 & 0.086 & 0.298 & 0.012 & 0.062 & 0.000 & 0.260 \\
\hspace{0.6em} ReSeek
& 0.501 & 0.684 & 0.409 & 0.436 & 0.288 & 0.157 & \underline{0.398} & \textbf{0.053} & 0.137 & 0.019 & 0.308 \\

\rowcolor{gray!15}
\hspace{0.6em} SARL & \underline{0.508} & \textbf{0.762} & \textbf{0.527} & \textbf{0.614}
& \textbf{0.645} & \textbf{0.272} & \textbf{0.548}
& \textbf{0.053} & \textbf{0.261} & \textbf{0.103} & \textbf{0.429} \\
\bottomrule
\end{tabularx}

\label{tables:main_results}
\end{table*}

\subsection{Experimental Setup}
\label{subsec:experimental_setup}

\paragraph{Datasets}
We evaluate on a broad set of in-domain and out-of-domain benchmarks spanning varying complexity. For single-hop question answering, we use Natural Questions (NQ)~\citep{kwiatkowski2019natural}, TriviaQA~\citep{joshi2017triviaqa}, and PopQA~\citep{mallen-etal-2023-trust}. For multi-hop reasoning, we use HotpotQA~\citep{yang2018hotpotqa}, 2WikiMultihopQA (2Wiki)~\citep{ho-etal-2020-constructing}, Musique~\citep{trivedi2022musique}, and Bamboogle~\citep{bamboogle}, sampling 1,000 examples per benchmark. For deep-research tasks requiring long-horizon search and synthesis, we evaluate on GAIA~\citep{mialon2023gaia}, FRAMES~\citep{krishna-etal-2025-frames}, and BrowseComp-Plus (BC-Plus)~\citep{chen2025browsecompplus}.
To account for multiple valid surface forms of an answer, we adopt an LLM-as-a-judge evaluation protocol~\citep{zheng2023judging} using GPT-4.1~\citep{openai2025gpt41} as the judge. All reported numbers are avg@4, averaged across four sampled completions per query to mitigate sampling stochasticity. We additionally report exact match (EM) and F1 scores in Appendix~\ref{app:result_em_f1}.

\paragraph{Baselines}
We compare our approach against several representative baselines spanning different paradigms of reasoning and retrieval. 
\textbf{(1) Without retrieval:} The model answers using only its parametric knowledge. We consider Direct Generation, supervised fine-tuning (SFT), and R1-style training (R1)~\citep{jin2025search}. 
\textbf{(2) With retrieval:} The model retrieves supporting documents during inference. We evaluate Naive RAG~\citep{lewis2020retrieval} and IRCoT~\citep{trivedi2023interleaving}. 
\textbf{(3) Search agents:} The model is equipped with retrieval tools and generates multi-step interaction trajectories. This category includes Search-R1~\citep{jin2025search} and ASearcher~\citep{asearcher}, trained to follow a standard ``think-search-answer'' paradigm, as well as AutoRefine~\citep{shi2025search} and ReSeek~\citep{li2025reseek}, additionally trained to incorporate refinement or judgment steps. 
All baselines are built on Qwen2.5-7B~\citep{yang2025qwen3}, except for Direct Generation and SFT (instruct variant), and are reproduced under the same experimental setting for fair comparison. 
For broader-paradigm reference, we additionally include published results from RAG-R1~\citep{tan2026rag}, a two-stage SFT+RL framework, and MA-RAG~\citep{marag2025}, a multi-agent framework with four specialized agents. As these methods rely on framework-specific pipelines or backbones, we do not reproduce them under our setup; full setup comparison and results are deferred to Appendix~\ref{appendix:cross_paradigm}.

\paragraph{Implementation Details}
All models are trained on the same 45K training set for one epoch using 8 B200 GPUs, with early termination upon divergence. For validation, we sample 500 examples from each dataset and select checkpoints based on the highest validation scores. We set the maximum number of interaction turns to 32 and sample 5 trajectories per query during training. For retrieval, we use e5-base-v2 over the December 2018 Wikipedia dump, returning top-$3$ documents per query, except for BrowseComp-Plus, where we use Qwen3-Embedding-8B with a custom curated Wikipedia dump for answer coverage. For reward design, we set $\alpha=0.8$ and $\lambda=0.5$. SARL is trained on the SkyRL framework~\citep{griggs2025skrylv01}; baselines use their original libraries. Training curves across all reproduced baselines and a training cost analysis are provided in Appendix~\ref{sec:appendix_training_details} and Appendix~\ref{appendix:training_cost}.

\begin{figure*}[t]
    \centering
    \includegraphics[width=\textwidth]{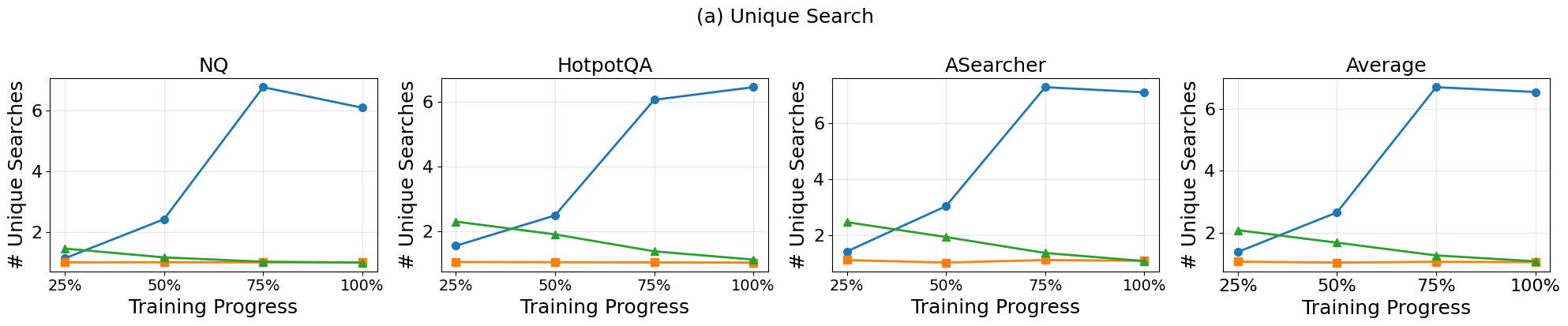}

    \vspace{6pt}

    \includegraphics[width=\textwidth]{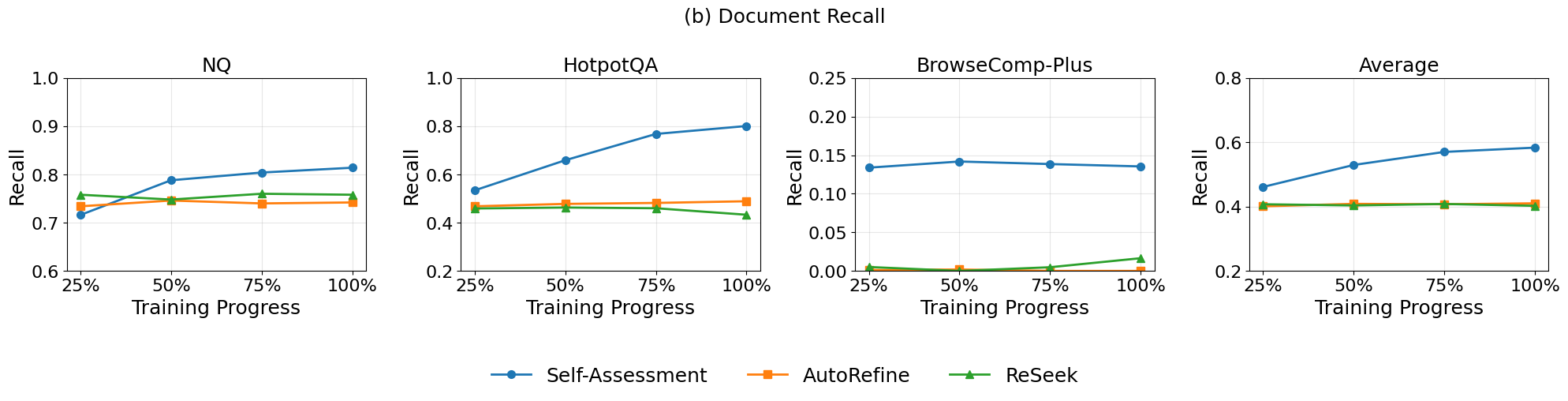}

\caption{
Search behavior of SARL compared with AutoRefine and ReSeek across training progress. Top: number of unique search queries across benchmarks. Bottom: document recall across benchmarks. SARL improves on both metrics over training, while baselines remain flat or decline.
}
    \label{fig:analysis_search_recall}
\end{figure*}

\subsection{Main Results}
\label{subsec:main_results}

Table~\ref{tables:main_results} presents the overall performance across our benchmark suite. SARL achieves an average score of 0.429, a 24.3\% relative improvement over the strongest baseline (ASearcher: 0.345). Despite only NQ and HotpotQA overlapping with our training data, SARL demonstrates consistent gains across all benchmarks, with peak improvements of 58.1\% on Musique and 48.3\% on FRAMES against the best per-benchmark baselines, suggesting that SARL learns generalizable search strategies that transfer to unseen task distributions.

We further compare performance gains across three task categories, measured against the strongest per-benchmark baseline within each category. While SARL achieves modest gains of 3.9\% on single-hop QA, the gains become substantially more pronounced on complex reasoning scenarios, with average gains of 35.9\% on multi-hop QA and 27.9\% on deep research tasks. This performance pattern indicates that SARL's self-assessment mechanism is particularly effective for tasks requiring multiple rounds of information retrieval. As tasks grow more complex and demand iterative refinement of search queries based on accumulated knowledge, the ability to assess what information has been obtained and what remains missing becomes increasingly critical for both efficiency and accuracy.

We further observe a previously unreported failure mode in existing methods under extended search horizons. With the maximum search turn limit increased to 32 and longer-trajectory synthetic datasets included in training, both AutoRefine and ReSeek exhibit severe training instability and performance degradation, with AutoRefine collapsing to fewer than one search turn on average and ReSeek's validation scores dropping toward zero late in training. This stems from an architectural choice: both methods generate reflection text inline at every turn, making them vulnerable to context growth and training collapse over long-horizon trajectories. SARL avoids this failure mode by decoupling the trajectory from the self-assessment step and applying fine-grained rewards to the assessment, enabling stable scaling to long search horizons (Figure~\ref{fig:analysis_search_recall}; mechanism and training curves in Appendix~\ref{sec:appendix_multiturn} and Appendix~\ref{sec:appendix_training_details}). We analyze reward modeling effects in Section~\ref{subsec:reward_ablation}.

For broader context, we additionally reference published results from RAG-R1, a two-stage SFT+RL pipeline, and MA-RAG, a training-free multi-agent framework. SARL achieves a 23.4\% relative improvement over RAG-R1 on the seven shared benchmarks (0.554 vs.\ 0.449 average), demonstrating that SARL outperforms pipelines that combine SFT warmup with RL. Against MA-RAG, SARL substantially outperforms the 8B variant on multi-hop reasoning (HotpotQA: 0.614 vs.\ 0.403; 2Wiki: 0.645 vs.\ 0.318) and remains ahead even of the 70B variant on these benchmarks (0.614 vs.\ 0.507; 0.645 vs.\ 0.431), indicating that internalizing search-state awareness through training can be more effective than scaling backbone size or inference-time orchestration. Setup details and per-benchmark results are provided in Appendix~\ref{appendix:cross_paradigm}.
\section{Analysis}
\label{section:analysis}

\begin{figure*}[t]
\centering
\includegraphics[width=0.9\textwidth]{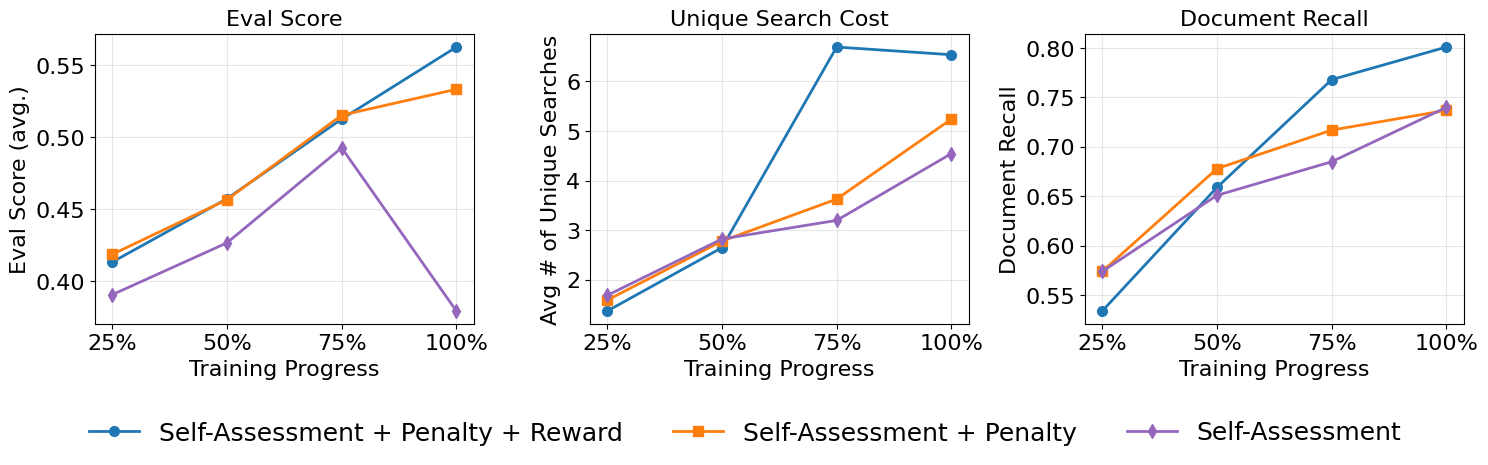}
\caption{
Ablation of SARL's reward components across training progress, reporting eval score, unique search cost, and document recall (left to right). Only the full configuration (Self-Assessment + Penalty + Reward) yields stable improvements.
}
\label{figure:analysis_ablation}
\end{figure*}

\subsection{Search Diversity and Evidence Recall}
\label{subsec:search_behavior}

A key factor underlying the effectiveness of search agents is their ability to actively explore missing information and retrieve relevant evidence. To examine how this behavior evolves during training, we analyze both the diversity of search queries and the relevance of retrieved documents at different training stages.

We first study the diversity of search behavior by measuring the number of unique search queries generated over training, which filters out repeated or redundant queries. As shown in Figure~\ref{fig:analysis_search_recall}(a), SARL exhibits a clear and consistent increase in the number of unique search queries as training progresses, reaching an average of over six distinct queries toward the end of training. This trend is highly consistent across NQ, HotpotQA, and ASearcher, and becomes more pronounced for datasets requiring more complex reasoning. These results indicate that SARL progressively learns to explore missing information more actively rather than prematurely terminating search. In contrast, AutoRefine and ReSeek show a declining trend in unique search queries over training, suggesting that existing RL-based methods fail to encourage sustained exploration and instead converge toward increasingly repetitive or shallow search behavior.

We next analyze the relevance of retrieved documents by tracking document recall throughout training. As illustrated in Figure~\ref{fig:analysis_search_recall}(b), SARL steadily improves recall across datasets, demonstrating its growing ability to retrieve relevant evidence before producing a final answer. This improvement is particularly notable on NQ and HotpotQA, where recall increases monotonically over training. Although the gains on BrowseComp-Plus are more modest, this dataset presents substantially higher difficulty, with target evidence being sparsely distributed and weakly aligned with surface-level queries. Importantly, SARL still maintains a consistent trend, indicating that it continues to search for relevant information even under challenging long-horizon settings. In contrast, AutoRefine and ReSeek exhibit largely static recall curves across all datasets, and perform especially poorly on BrowseComp-Plus, where they rarely retrieve relevant evidence. This suggests that their training objectives do not generalize well to complex tasks that require sustained, multi-step information seeking.

The improvements in both search diversity and evidence recall stem from SARL's ability to recognize its own search state. In Appendix~\ref{subsec:preliminary_results}, we directly evaluate this capability by analyzing the agent's reasoning trajectories, finding that SARL exhibits notably higher search-state awareness than existing baselines. This connection reinforces our central premise: by learning to recognize when retrieved evidence is insufficient, the model is naturally driven to issue more diverse queries and recover more relevant evidence.

\subsection{Effect of Assessment Penalty and Reward}
\label{subsec:reward_ablation}

We conduct an ablation study to isolate the contributions of the key components of SARL. We compare three variants: (1) Self-Assessment only, (2) Self-Assessment with assessment penalty, and (3) full SARL with both assessment penalty and reward. Figure~\ref{figure:analysis_ablation} reports validation performance, unique search cost, and document recall across training progress.

As shown in Figure~\ref{figure:analysis_ablation}, incorporating both the assessment penalty and the assessment reward yields the strongest and most stable improvements across all metrics. In contrast, training with self-assessment alone leads to instability, with performance peaking mid-training and then sharply degrading toward the end. This behavior suggests that self-assessment signals by themselves are insufficient to guide effective long-horizon search, and may even encourage premature termination or degenerate search behavior when not properly constrained.

Introducing an assessment penalty substantially mitigates this issue, indicating that penalizing reliance on assessments plays a critical role in encouraging the model to construct complete search trajectories prior to negative assessments. Furthermore, the assessment reward provides a complementary learning signal that directly incentivizes productive use of self-assessment. As shown in the number of unique searches and recall, the full SARL model exhibits a marked improvement in both unique search queries and document recall at later training stages, with over a twofold increase in unique searches compared to the self-assessment-only variant. These gains translate into consistent improvements in final evaluation performance. Together, these results demonstrate that fine-grained rewards based on how the model's own assessments affect search outcomes are essential for encouraging sustained exploration of missing information and for learning effective search strategies.

Beyond establishing that both components are necessary, we further study the sensitivity of SARL to the assessment penalty coefficient $\alpha$ and the assessment reward coefficient $\lambda$, with full results provided in Appendix~\ref{appendix:hyperparam}. With $\lambda=0.5$ fixed, performance is robust across a wide range of $\alpha \in \{0.2, 0.5, 0.8\}$, with final validation scores spanning a narrow 4-point range (51.7--55.5). Varying $\lambda$ with $\alpha=0.8$ fixed reveals a more pronounced asymmetry: $\lambda=0.2$ briefly reaches a competitive best score (55.5) but collapses to 42.6 by the end of training, while $\lambda=0.5$ and $\lambda=0.8$ both remain stable. This pattern indicates that under-weighting the assessment reward leads to training instability rather than under-fitting; the model can transiently learn to assess search state but lacks sufficient incentive to leverage assessments productively, eventually drifting back toward shallow search behavior. Adequate weighting of the assessment reward is therefore essential for sustaining the gains observed under our default configuration ($\alpha=0.8, \lambda=0.5$).

\subsection{Generalizability across Model Scale and Family}
\label{subsec:generalizability}

To assess whether SARL's gains transfer beyond the Qwen2.5-7B backbone, we evaluate two additional configurations: Llama-3.1-8B-Instruct (different model family, similar scale) and Qwen2.5-14B (same family, larger scale). For both, we report validation scores at training steps 100, 200, and 300, evaluated on 500 held-out examples per dataset from NQ, HotpotQA, and ASearcher. Per-step results are provided in Appendix~\ref{appendix:cross_model}.

\paragraph{Cross-family generalization.}
SARL transfers effectively to a different model family. While GRPO-trained Llama-3.1-8B-Instruct collapses over training (52.3 $\rightarrow$ 51.3 $\rightarrow$ 44.4), SARL exhibits the opposite trajectory, with its average score increasing steadily from 51.2 to 54.3. By step 300, SARL outperforms GRPO on every benchmark, with a +9.9 gap in average score. This contrast illustrates that the assessment-driven training signal stabilizes long-horizon RL training across model families, whereas standard outcome-based RL is prone to performance collapse on a different backbone.

\paragraph{Scale generalization.}
We further compare SARL with the same Qwen2.5 backbone at 7B and 14B parameters. The 14B variant outperforms the 7B variant at every training step and on every benchmark. Although the gap narrows from +6.4 average points at step 100 to +3.6 points at step 300, the 14B variant maintains a consistent advantage throughout training. The improvement is most pronounced on ASearcher (+5.6 at step 300), which involves the longest search trajectories among the three benchmarks.

\section{Related Work}
\label{section:related_work}

\paragraph{Retrieval-Augmented Generation}
Retrieval-augmented generation (RAG) mitigates the inherent limitations of LLMs by conditioning text generation on externally retrieved evidence~\citep{lewis2020retrieval, guu2020retrieval}. While standard pipelines follow a single-step retrieve-then-generate flow, recent advances introduce multi-document aggregation~\citep{izacard-grave-2021-leveraging} and \emph{self-controlled} approaches that decide when to retrieve and reflect on retrieved evidence, via prompted reasoning traces~\citep{yao2022react}, supervised reflection tokens~\citep{asai2024selfrag}, or learned tool-use policies~\citep{schick2023toolformer, nakano2021webgpt}. These methods improve factual grounding when relevant evidence is accessible, but static or shallowly iterative retrieval often becomes brittle when evidence is scattered across many sources or requires sustained clarification~\citep{yang2018hotpotqa, trivedi2022musique, trivedi2023interleaving}.

\paragraph{Internalizing Search-State Awareness through RL}
Search agents formulate information seeking as a sequential decision problem, where the model iteratively decides when to search, what to query, and when to conclude over long interaction trajectories. Recent works increasingly use RL to optimize these policies based on outcome feedback, as in Search-R1~\citep{jin2025search} and related approaches~\citep{Search-o1, asearcher}. Beyond short-horizon QA, growing interest in ``deep research'' behaviors has motivated large-scale training recipes for long-horizon web interaction~\citep{tongyi2025deepresearch, qiao2025webresearcher, tao2025webleaper, liu2025webexplorer, lu2025deepdive}. However, training on final outcomes often fails to ensure reliable trajectories: agents may reach correct answers through non-faithful intermediate reasoning, premature termination, or unsupported assumptions, a form of reward hacking and weak credit assignment under sparse final rewards~\citep{xu2025beyond, zhang2026carr, yu2026webanchor}.

To mitigate reward sparsity, prior work introduces SFT-warmup pipelines such as RAG-R1~\citep{tan2026rag}, which bootstrap tool use through curated or synthesized trajectories before applying RL, and multi-agent frameworks such as MA-RAG~\citep{marag2025}, which delegate planning, retrieval, and synthesis to specialized agents at inference time. Other lines introduce intermediate or process rewards to encourage auxiliary behaviors such as refinement, evidence quality, or faithfulness~\citep{shi2025search, li2025reseek}, or improve long-horizon optimization through planning anchors, tree-based rollouts~\citep{yu2026webanchor, ji2025treegrpo}, and explicit context or memory management~\citep{sun2025contextfolding, qian2026memobrain, li2025webweaver}.

In contrast, SARL treats self-assessment as an \emph{internal} training signal within a single policy, learned through outcome-based credit assignment rather than prompting heuristics or supervised reflection demonstrations. SARL rewards measurable changes that follow from an assessment step (e.g., correcting an incorrect answer) and penalizes excessive reliance on assessments, encouraging trajectory repair when necessary while progressively internalizing knowledge-aware search behavior. This design enables accurate, efficient search to emerge from RL alone, without additional long-horizon data, auxiliary judges, or inference-time orchestration.
\section{Conclusion}
\label{section:conclusion}
We introduce Self-Assessment Reinforcement Learning (SARL), a framework that internalizes search-state awareness within retrieval-augmented LLMs through a reward design that balances corrective assessment rewards with efficiency penalties. Across ten benchmarks, SARL consistently outperforms baselines under fixed training conditions, with particularly strong gains on complex tasks: 35.9\% on multi-hop QA and 27.9\% on deep research benchmarks. These improvements stem from the model's progressively internalized ability to identify information gaps and refine retrieval, reflected in a twofold increase in unique search queries and notably higher document recall. By shifting search-state awareness from a token-level prompted process into an internalized reasoning capacity, SARL offers a path toward knowledge-aware agents that self-correct without reliance on external feedback at inference time.

\bibliography{example_paper}
\bibliographystyle{icml2026}

\newpage
\appendix
\onecolumn
\section{Appendix}

\subsection{Limitations}
\label{appendix:limitations}

While SARL demonstrates effective performance across a broad range of search-intensive tasks, several limitations remain.

\paragraph{Reliance on base model capability.}
SARL depends on the base model's ability to recognize when retrieved evidence is sufficient to support a candidate answer, an assumption that does not hold universally. Our central premise experiments (Appendix~\ref{appendix:central_premise}) demonstrate that this judgment is reliable enough to provide an effective training signal at the 7B scale, allowing SARL to retain both efficiency and effectiveness without incurring the overhead of an external evaluator model as in RLVR-style methods. Investigating whether incorporating a more powerful external model can yield further gains is a promising direction for future work.

\paragraph{Focus on search-intensive tasks.}
Our evaluation focuses exclusively on search-augmented question answering. While this scope is among the broadest within the search agent literature, demonstrating effectiveness on long-horizon deep research benchmarks even at the 7B scale, SARL's applicability to other domains such as code generation, mathematical reasoning, or general agentic planning remains an open question.

\paragraph{Compatibility with broader pipelines.}
We present SARL as a standalone RL framework to isolate the contribution of self-assessment as a training signal. Whether SARL combines synergistically with broader pipelines, such as SFT-warmup or multi-agent orchestration, remains to be explored, and we leave this investigation to future work.

\subsection{Example Trajectories of Search-State Awareness}
\label{appendix:trajectory_examples}

To illustrate the failure modes discussed in Section~\ref{section:introduction} and the search-state awareness behavior promoted by SARL, we provide representative trajectories from our evaluation set. A complementary empirical comparison across all baselines is provided in Appendix~\ref{subsec:preliminary_results}. For readability, retrieved documents are abbreviated to the relevant excerpt; we preserve the model outputs verbatim, including any malformed tags as produced by the model.

\paragraph{Hallucinated reasoning (Search-R1).}
The agent explicitly recognizes that the retrieved evidence does not satisfy the question's constraint, yet commits to an unsupported answer.

\begin{lstlisting}[
  basicstyle=\ttfamily\footnotesize,
  breaklines=true,
  breakatwhitespace=true,
  columns=fullflexible
]
Q: What political office was held by two successive occupants of the same castle located in Lauderdale, Scotland?
GT: Keeper of the Privy Seal of Scotland

<think> First, I will search for information:
<search> Who were the political occupants of the same castle in Lauderdale, Scotland? </search>

<information> ... Thirlestane Castle has been the seat of the Earls of Lauderdale ... </information>

<think> From the provided information, Thirlestane Castle has been the seat of the Earls of Lauderdale.
<think] However, the question asks for a political office held by two successive occupants of the same castle.
<think> None of the provided information directly states a political office held by two successive occupants of Thirlestane Castle.
<search> What political office did the Earls of Lauderdale hold? </search>

<information> ... John Maitland, 1st Earl of Lauderdale, was President of the Parliament of Scotland ... Frederick Maitland, 14th Earl of Lauderdale, was a Representative Peer for Scotland ... </information>

<think> These are political offices, but they are not held by "successive occupants" in the same castle.
Without further information, I cannot provide a specific answer to the question. However, if we consider the Earls of Lauderdale as "occupants" of Thirlestane Castle, then the answer would be:
<answer> Earl of Lauderdale </answer>
\end{lstlisting}

The agent's reasoning explicitly acknowledges that no political office held by successive occupants has been identified, yet the agent proceeds to commit to an answer regardless. The final response also confuses an aristocratic title (Earl of Lauderdale) with a political office, indicating a failure to internalize what evidence the trajectory has actually established.

\paragraph{Prematurely terminated search (AutoRefine).}
The agent commits to an answer after a single search, with the refinement token operating as a structural marker rather than a substantive reflection on retrieval state.

\begin{lstlisting}[
  basicstyle=\ttfamily\footnotesize,
  breaklines=true,
  breakatwhitespace=true,
  columns=fullflexible
]
Q: Who is Catherine I, Latin Empress's father-in-law?
GT: Philip III of France

<search> Catherine I, Latin Empress's father-in-law </search>

<information> ... [retrieved documents on Catherine I and the Courtenay family] ... </information>

<refine> Based on the documents, Catherine I, Latin Empress's father-in-law is Philip I of Courtenay. </refine>

<answer> Philip I of Courtenay </answer>
\end{lstlisting}

The agent issues a single search and immediately commits to an answer, despite the multi-step nature of the question (identifying Catherine I's husband, then the husband's father). The \texttt{<refine>} token contains an unverified assertion rather than a reflection on what the retrieved documents actually support, illustrating how structural reflection tokens fail to enforce evidence-grounded reasoning when not tied to a substantive learning signal.

\paragraph{Successful search-state awareness (SARL).}
The agent decomposes a multi-hop question into sequential subgoals, identifies the constraints satisfied at each step, and issues targeted follow-up searches to recover the remaining evidence.

\begin{lstlisting}[
  basicstyle=\ttfamily\footnotesize,
  breaklines=true,
  breakatwhitespace=true,
  columns=fullflexible
]
Q: What is the date of birth of Emilie Hegh Arntzen's mother?
GT: 27 April 1960

<think> To answer this question, we'll first need to gather information on Emilie Hegh Arntzen and her mother. We'll break this task down into smaller parts:
1. Identify Emilie Hegh Arntzen.
2. Find information on her mother.
3. Determine the date of birth of her mother.
<search> Who is Emilie Hegh Arntzen? </search>

<information> ... Emilie Hegh Arntzen is a Norwegian handball player ... daughter of Hanne Hegh ... </information>

<think> Now, we have enough information to focus on Emilie Hegh Arntzen's mother, Hanne Hegh.
<search> Who is Hanne Hegh, the mother of Emilie Hegh Arntzen? </search>

<information> ... Hanne Hegh is a Norwegian handball player ... </information>

<think> We have identified the correct Hanne Hegh as Emilie Hegh Arntzen's mother, who is a Norwegian handball player. We need to find her date of birth.
<search> When was Hanne Hegh, the Norwegian handball player, born? </search>

<information> ... Hanne Hegh ... born on 27 April 1960 ... </information>

<think> We have identified the correct person, Hanne Hegh, who is Emilie Hegh Arntzen's mother and a Norwegian handball player. We've also found the date of her birth, which is 27 April 1960.
<answer> 27 April 1960 </answer>
\end{lstlisting}

The agent explicitly tracks search-state awareness throughout the trajectory: at each step, it states what has been resolved (\textit{``Now, we have enough information to focus on... mother''}, \textit{``We have identified the correct Hanne Hegh''}) and what remains missing (\textit{``We need to find her date of birth''}) before issuing a targeted follow-up search. This contrasts with the failure modes above: the agent neither commits to an unsupported answer nor terminates after a single search, but instead progressively decomposes the question and verifies each subgoal.


\subsection{Validating the Central Premise of SARL}
\label{appendix:central_premise}

The motivation for SARL rests on the premise that models can meaningfully identify and correct their own erroneous trajectories when observed from a separate context. We empirically validate this premise through two controlled experiments.

\paragraph{Experiment 1: Foundational capability in base models.}
We first verify that the base model possesses the foundational capability for self-assessment before any SARL training. We collect incorrect (and therefore incomplete) trajectories generated by Search-R1 on our validation set, and prompt Qwen2.5-7B with the self-assessment prompt (Appendix~\ref{appendix:assessment_prompt}). The base model correctly identifies \textbf{73.2\%} of flawed trajectories as \texttt{INCOMPLETE}, demonstrating that the capability exists prior to training and providing a meaningful starting point for SARL to build upon.

\paragraph{Experiment 2: Meaningfulness of the self-assessment signal.}
To determine whether SARL's improvement stems from genuine error identification rather than spurious factors (e.g., simply allowing extra search turns), we replace the model's self-assessment with randomized feedback (\textit{``The search process is \texttt{INCOMPLETE}/\texttt{COMPLETE}''}) under otherwise identical training conditions.

\begin{table}[h]
\centering
\small
\begin{tabular}{cl|cccc}
\toprule
Step & Variant & NQ & HotpotQA & ASearcher & Avg \\
\midrule
\multirow{2}{*}{100} & Random & 40.8 & 52.4 & 39.8 & 44.3 \\
                     & SARL   & 44.0 & 52.2 & 41.0 & 45.7 \\
\midrule
\multirow{2}{*}{200} & Random & 44.8 & 52.8 & 42.2 & 46.6 \\
                     & SARL   & 49.2 & 56.0 & 48.6 & 51.3 \\
\midrule
\multirow{2}{*}{339} & Random & 36.4 & 48.6 & 38.6 & 41.2 \\
                     & SARL   & 52.4 & 61.8 & 52.2 & 55.5 \\
\bottomrule
\end{tabular}
\caption{SARL vs. randomized assessment across training stages. SARL consistently outperforms the random baseline, with the gap widening over training.}
\label{table:randomized_assessment}
\end{table}

SARL outperforms random assessment across all training stages, with the gap widening as training progresses. By the end of one epoch (step 339), SARL reaches \textbf{55.5} average accuracy while random assessment \textit{degrades} to 41.2. Notably, the random variant exhibits training instability, degrading from 46.6 (step 200) to 41.2 (step 339), suggesting that arbitrary assessment signals introduce noise that destabilizes training over time. This confirms that the content of self-assessment, not merely the structural opportunity for extra search turns, drives SARL's improvement.

Together, these experiments validate two foundations of SARL's design. First, the foundational capability for self-assessment exists in base models, providing a meaningful starting point for SARL to amplify rather than fabricate. Second, the assessment signal SARL leverages is genuinely meaningful: arbitrary feedback under otherwise identical conditions not only fails to improve performance but actively destabilizes training, indicating that the content of the assessment carries real corrective information. These results empirically support the central premise stated in our introduction: search-state awareness can be cultivated through reinforcement learning when the underlying model has a sufficient starting capability and the training signal carries genuine corrective information.


\subsection{Performance Evaluation on EM and F1 Metrics}
\label{app:result_em_f1}

\begin{table*}[h]
\centering
\caption{Performance comparison on single-hop QA, multi-hop QA, and deep research benchmarks. The results are presented in the format of \textbf{Exact Match (EM) / F1}.}
\label{tab:main_table_em_f1}
\resizebox{\textwidth}{!}{%
\begin{tabular}{lcccccccccc}
\toprule
\multirow{2}{*}{\textbf{Methods}} & \multicolumn{3}{c}{\textbf{Single-Hop QA}} & \multicolumn{4}{c}{\textbf{Multi-Hop QA}} & \multicolumn{3}{c}{\textbf{Deep Research}} \\
\cmidrule(lr){2-4} \cmidrule(lr){5-8} \cmidrule(lr){9-11}
 & \textbf{NQ} & \textbf{TriviaQA} & \textbf{PopQA} & \textbf{HotpotQA} & \textbf{2Wiki} & \textbf{Musique} & \textbf{Bamboogle} & \textbf{GAIA} & \textbf{FRAMES} & \textbf{BC-Plus} \\

\midrule
\multicolumn{11}{l}{\textit{w/o Retrieval}} \\
\hspace{3mm} Direct Generation & 0.051 / 0.166 & 0.319 / 0.415 & 0.160 / 0.160 & 0.160 / 0.255 & 0.193 / 0.248 & 0.016 / 0.096 & 0.066 / 0.152 & 0.015 / 0.074 & 0.029 / 0.128 & 0.002 / 0.025 \\
\hspace{3mm} SFT & 0.168 / 0.239 & 0.355 / 0.415 & 0.150 / 0.181 & 0.163 / 0.233 & 0.185 / 0.228 & 0.026 / 0.092 & 0.114 / 0.157 & 0.044 / 0.095 & 0.036 / 0.076 & 0.002 / 0.032 \\
\hspace{3mm} R1 & 0.217 / 0.295 & 0.429 / 0.498 & 0.188 / 0.230 & 0.209 / 0.286 & 0.277 / 0.308 & 0.050 / 0.119 & 0.200 / 0.270 & 0.046 / 0.090 & 0.121 / 0.122 & 0.006 / 0.032 \\
\midrule
\multicolumn{11}{l}{\textit{w/ Retrieval}} \\
\hspace{3mm} Naive RAG & 0.256 / 0.375 & 0.404 / 0.495 & 0.332 / 0.408 & 0.195 / 0.294 & 0.189 / 0.236 & 0.037 / 0.097 & 0.070 / 0.140 & 0.012 / 0.033 & 0.037 / 0.088 & 0.013 / 0.031 \\
\hspace{3mm} IRCoT & 0.290 / 0.407 & 0.434 / 0.541 & 0.392 / 0.471 & 0.281 / 0.395 & 0.303 / 0.364 & 0.069 / 0.138 & 0.192 / 0.278 & 0.036 / 0.071 & 0.074 / 0.147 & 0.039 / 0.068 \\
\midrule
\multicolumn{11}{l}{\textit{Search Agents}} \\
\hspace{3mm} Search-R1 & 0.449 / 0.522 & 0.548 / 0.637 & 0.469 / 0.513 & 0.355 / 0.463 & 0.393 / 0.442 & 0.117 / 0.202 & 0.288 / 0.388 & 0.032 / 0.054 & 0.100 / 0.160 & 0.056 / 0.083 \\
\hspace{3mm} ASearcher & 0.395 / 0.510 & 0.506 / 0.616 & 0.418 / 0.492 & 0.359 / 0.495 & 0.302 / 0.373 & 0.106 / 0.197 & 0.226 / 0.361 & 0.034 / 0.081 & 0.100 / 0.195 & 0.077 / 0.117 \\
\hspace{3mm} AutoRefine & 0.471 / 0.551 & 0.504 / 0.587 & 0.429 / 0.461 & 0.242 / 0.335 & 0.153 / 0.183 & 0.058 / 0.096 & 0.224 / 0.313 & 0.010 / 0.010 & 0.038 / 0.063 & 0.000 / 0.000 \\
\hspace{3mm} ReSeek & 0.109 / 0.250 & 0.280 / 0.400 & 0.157 / 0.274 & 0.120 / 0.251 & 0.082 / 0.182 & 0.041 / 0.108 & 0.108 / 0.215 & 0.044 / 0.063 & 0.038 / 0.102 & 0.011 / 0.020 \\
\hspace{3mm} SARL & \textbf{0.506 / 0.577} & \textbf{0.583 / 0.677} & \textbf{0.505 / 0.545} & \textbf{0.448 / 0.588} & \textbf{0.537 / 0.619} & \textbf{0.198 / 0.315} & \textbf{0.402 / 0.521} & \textbf{0.049 / 0.094} & \textbf{0.161 / 0.240} & \textbf{0.086 / 0.128} \\
\bottomrule
\end{tabular}%
}
\end{table*}

Table~\ref{tab:main_table_em_f1} presents the EM and F1 scores for the main results. Consistent with the main evaluation results, SARL demonstrates consistently superior performance compared to other methods across all benchmarks in terms of EM and F1 metrics.

\subsection{Overview of Training Stability and Performance}
\label{sec:appendix_training_details}
\begin{figure*}[t]
\centering
\includegraphics[width=0.8\textwidth]{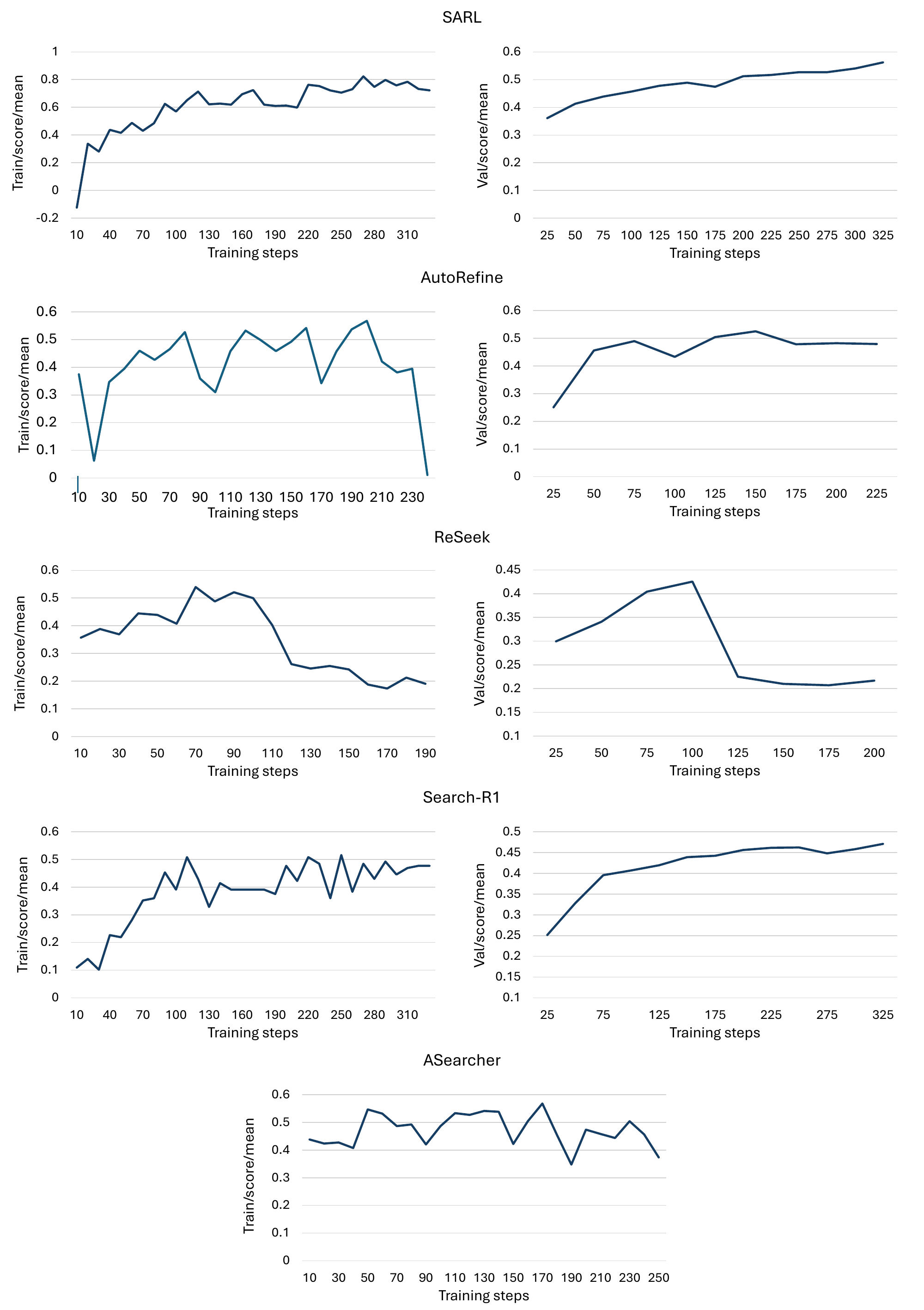}
\caption{
The training logs reveal a contrast in stability and learning efficacy between SARL and existing baselines. SARL achieves a stable, monotonic increase in both training and validation scores, reaching a peak validation score of approximately 0.56 by 325 steps. This indicates that its dual reward-penalty mechanism successfully internalizes search-state awareness without the model becoming over-reliant on explicit feedback.
}
\label{figure:appendix_training_log}
\end{figure*}
Figure~\ref{figure:appendix_training_log} depicts the training (per 10 step) / validation (per 25 step) logs across all models.

SARL demonstrates the most stable and consistent improvement across both training and validation curves.
The validation curve shows a steady upward trajectory throughout the 325 training steps, indicating that the model effectively internalizes search-state awareness without overfitting or collapsing.

Search-R1~\citep{jin2025search} and ASearcher~\citep{asearcher} models show a general upward trend in performance, but with noticeably more fluctuations in their training scores compared to SARL. While they reach competitive performance levels, they do not match the peak validation scores achieved by the SARL framework.

AutoRefine~\citep{shi2025search} and ReSeek~\citep{li2025reseek} models exhibit significant training instability. AutoRefine's performance peaks mid-training before sharply declining. ReSeek experiences a severe training collapse toward the end of its cycle, where scores drop toward zero, likely due to the difficulty of handling long-horizon trajectories.

\subsection{Training Cost Analysis}
\label{appendix:training_cost}
\begin{table}[h]
\centering
\small
\renewcommand{\arraystretch}{1.1}
\caption{Training cost comparison between SARL and Search-R1.}
\label{tab:training_cost}
\begin{tabular}{l c c c}
\toprule
& \textbf{Search-R1} & \textbf{SARL} & \textbf{Ratio} \\
\midrule
Total training time & 22.1 hrs & 51.3 hrs & 2.3$\times$ \\
Avg.\ time / step & 235 s & 544 s & 2.3$\times$ \\
Avg.\ trajectory length (tokens) & 1{,}742 & 4{,}393 & 2.5$\times$ \\
Avg.\ performance & 0.342 & 0.429 & $+25.4\%$ \\
\bottomrule
\end{tabular}
\end{table}
We provide a detailed comparison of training cost between SARL and Search-R1, both implemented within the SkyRL framework under identical training settings. We focus on this pairing because shared infrastructure isolates the cost difference attributable to the SARL training procedure itself, rather than confounding factors from differing RL libraries.
 While SARL incurs additional compute during training, invoking self-assessment until the model produces a ``COMPLETE'' judgment, the exponential penalty ($\alpha=0.8$) bounds these iterations in practice, discouraging repeated assessments and encouraging thorough searching upfront. At test time, self-assessment is not used; SARL-trained models follow the same think-search-answer pipeline as standard search agent baselines. We reproduce both methods under the same RL library and settings for the comparison below.

To quantify the training-time overhead, we compare SARL with Search-R1, reproduced under the same RL library and settings.
As shown in Table~\ref{tab:training_cost}, the close alignment between the trajectory length ratio (2.5$\times$) and the time ratio (2.3$\times$) indicates that the overhead stems primarily from the longer, more thorough search trajectories the model learns to produce, rather than from the self-assessment calls themselves. The resulting 2.3$\times$ training cost yields a 25.4\% relative improvement in average performance, representing a favorable cost-performance tradeoff. As discussed in Section~\ref{section:analysis}, these longer trajectories translate into significantly more unique search queries and higher document recall, indicating that the additional compute is invested in productive search behavior rather than wasted computation.

\subsection{Comparison with Broader-Paradigm Methods}
\label{appendix:cross_paradigm}

We provide setup details and per-benchmark scores for the broader-paradigm comparison referenced in Section~\ref{subsec:main_results}.

\paragraph{RAG-R1.}
RAG-R1 combines supervised fine-tuning on Qwen2.5-72B-Instruct-generated trajectories with GRPO-based RL, optionally extended with multi-query parallelism (RAG-R1-mq). It uses the instruct variant of our backbone model (Qwen2.5-7B-Instruct), a later version of our search corpus (2019 Wikipedia dump), a larger retriever (BGE-large-en-v1.5~\citep{bge_embedding}), and the same evaluation metric.

\paragraph{MA-RAG.}
MA-RAG orchestrates four agents (Planner, Step Definer, Extractor, QA) through chain-of-thought prompting, without any model fine-tuning. It uses the Llama3-Instruct family (8B, 70B), the same search corpus, a larger retriever (gte-multilingual-base~\citep{zhang2024mgte}), and the same evaluation metric. Note that MA-RAG uses the KILT~\citep{petroni-etal-2021-kilt} subsets of NQ, TriviaQA, HotpotQA, and 2WikiMultiHopQA for evaluation.

\paragraph{Discussion.}
On the seven shared benchmarks with RAG-R1, SARL achieves an average of 0.554 versus 0.449 for RAG-R1-sq and 0.457 for RAG-R1-mq, despite using only RL training without the SFT warmup that RAG-R1 relies on. The gap is most pronounced on multi-hop reasoning (HotpotQA: +0.122 over RAG-R1-sq, +0.119 over RAG-R1-mq; 2Wiki: +0.125 / +0.082; Musique: +0.085 / +0.080), and persists on out-of-domain benchmarks such as Bamboogle (+0.108 / +0.108) where RAG-R1 has no training exposure. This pattern indicates that the gains stem not from data overlap but from how the training signal shapes search behavior across complex queries.

Against MA-RAG, SARL substantially outperforms the 8B variant on multi-hop benchmarks and remains ahead even of the 70B variant on these same benchmarks (HotpotQA: 0.614 vs. 0.507; 2Wiki: 0.645 vs. 0.431), despite a 10$\times$ smaller backbone and no inference-time agent orchestration. On TriviaQA, MA-RAG (Llama3-8B) achieves 0.826 versus our 0.762. This gap is consistent with backbone parametric knowledge differences: without retrieval, Llama3-8B-Instruct scores 0.707 on TriviaQA versus 0.523 for Qwen2.5-7B-Instruct, indicating that the TriviaQA gap reflects the underlying model's factual recall rather than search capability. Across the remaining benchmarks, SARL's gains over both methods are most pronounced on multi-hop reasoning, consistent with the diagnostic findings in Section~\ref{subsec:main_results} that self-assessment is most beneficial when iterative refinement of search queries is required.

\begin{table}[h]
\centering
\small
\renewcommand{\arraystretch}{1.1}
\caption{Cross-paradigm comparison against RAG-R1~\citep{tan2026rag} and MA-RAG~\citep{marag2025}. ``--'' indicates the benchmark was not reported. Avg.\ is computed over the seven shared benchmarks.}
\label{tab:cross_paradigm}
\resizebox{\columnwidth}{!}{%
\begin{tabular}{l c c c c c c c c}
\toprule
\textbf{Method} & \textbf{NQ} & \textbf{TriviaQA} & \textbf{PopQA} & \textbf{HotpotQA} & \textbf{2Wiki} & \textbf{Musique} & \textbf{Bamboogle} & \textbf{Avg.} \\
\midrule
RAG-R1-sq (Qwen2.5-7B-Inst) & 0.429 & 0.599 & 0.477 & 0.492 & 0.520 & 0.187 & 0.440 & 0.449 \\
RAG-R1-mq (Qwen2.5-7B-Inst) & 0.423 & 0.608 & 0.479 & 0.495 & 0.563 & 0.192 & 0.440 & 0.457 \\
\midrule
MA-RAG (Llama3-8B-Inst) & 0.525 & 0.826 & -- & 0.403 & 0.318 & -- & -- & -- \\
MA-RAG (Llama3-70B-Inst) & 0.581 & 0.854 & -- & 0.507 & 0.431 & -- & -- & -- \\
\midrule
\rowcolor{gray!15}
SARL (Qwen2.5-7B, Ours) & 0.508 & 0.762 & 0.527 & 0.614 & 0.645 & 0.272 & 0.548 & \textbf{0.554} \\
\bottomrule
\end{tabular}%
}
\end{table}

\subsection{Analysis of Multi-Turn Efficiency}
\label{sec:appendix_multiturn}
\begin{table}[h]
\centering
\caption{Average number of search operations performed by different agents across multi-hop QA and deep research benchmarks. Results reflect the agents' ability to sustain information seeking over long horizons.}
{\resizebox{.8\textwidth}{!}{
\begin{tabular}{lccccccc}
\toprule
\textbf{Models}     & \textbf{HotpotQA} & \textbf{2Wiki} & \textbf{Musique} & \textbf{Bamboogle} & \textbf{GAIA} & \textbf{FRAMES} & \textbf{BC-Plus} \\ \midrule
\textbf{Search-R1}~\citep{jin2025search}  & 3.24              & 4.34           & 5.77             & 3.88               & 12.14         & 8.50            & 5.90                     \\ 
\textbf{ASearcher}~\citep{asearcher}  & 3.58              & 3.32           & 4.07             & 3.70               & 5.77          & 5.26            & 5.05                     \\ 
\textbf{AutoRefine}~\citep{shi2025search} & 0.79              & 0.79           & 0.55             & 0.90               & 0.12          & 0.42            & 0.01                     \\ 
\textbf{ReSeek}~\citep{li2025reseek}     & 8.78              & 13.79          & 16.01            & 8.52               & 21.90         & 15.81           & 10.50                    \\ 
\textbf{SARL (\textsc{Ours})}       & 7.94              & 8.62           & 9.16             & 8.16               & 11.28         & 10.13           & 8.23                     \\ \bottomrule
\end{tabular}
}}{}
\label{tables:search_counts}
\end{table}
Building on the training instability observed in Appendix~\ref{sec:appendix_training_details}, we further analyze the search behavior driving these failures.

As shown in Table~\ref{tables:search_counts}, SARL demonstrates a robust and sustained exploration strategy across both multi-hop reasoning and deep research benchmarks.
SARL maintains a high frequency of search operations, averaging between 8.16 and 11.28 turns for complex tasks. This indicates that the model has successfully learned to fill information gaps through iterative retrieval rather than prematurely terminating the search.
While ReSeek~\citep{li2025reseek} exhibits the highest raw search counts, often exceeding 20 turns, its performance is hampered by training instability. In contrast, SARL achieves a superior balance, matching or exceeding the search depth of Search-R1~\citep{jin2025search} and ASearcher~\citep{asearcher} while maintaining significantly higher accuracy.

A notable observation in Table~\ref{tables:search_counts} is the exceptionally low search count for AutoRefine~\citep{shi2025search}, which fails to exceed 0.90 turns on average across all datasets. This phenomenon is not merely a matter of efficiency but a symptom of training collapse. Under extended search horizons (up to 32 turns), AutoRefine exhibits significant instability. The model tends to converge toward a single turn because it suffers from Lazy Likelihood-Displacement (LLD), a systematic reduction in the likelihood of both correct and incorrect responses during GRPO~\citep{deng2025grpo}. Because AutoRefine requires the model to generate explicit reflection text at every turn, it is highly susceptible to the ``LLD Death Spiral", where declining likelihoods lead the model to give up on complex trajectories, resulting in premature and often hallucinated conclusions.


\subsection{Evaluation of Search-State Awareness}
\label{subsec:preliminary_results}

\begin{figure}[h]
\centering
\includegraphics[width=0.8\columnwidth]{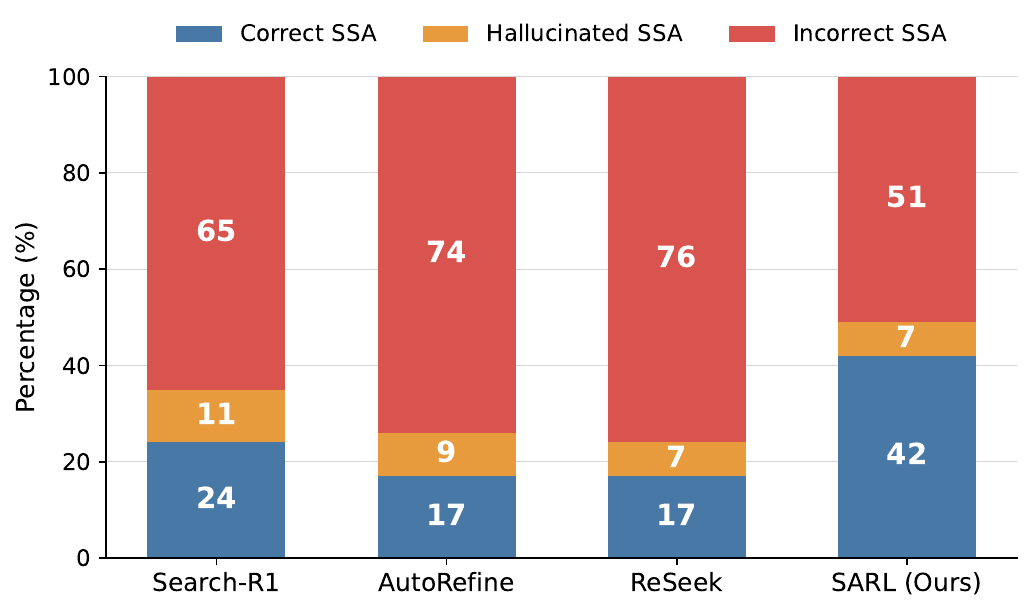}
\caption{
Comparison of Search-State Awareness (SSA) levels across SARL and baseline frameworks. SARL achieves a Correct SSA of 42\%, compared to 24\% for Search-R1 and 17\% for AutoRefine and ReSeek.
}
\label{figure:ka_barplot}
\end{figure}

In Section~\ref{section:introduction}, we identified search-state awareness as a core capability missing in current RL-based search agents, motivating two failure modes we observe in practice: hallucinated reasoning and prematurely terminated search. Here we directly evaluate whether SARL improves search-state awareness relative to existing baselines, providing the empirical comparison referenced in the introduction.

We sample 100 reasoning trajectories from HotpotQA, 2WikiMultiHopQA, Musique, and Bamboogle, and use GPT-5.2~\citep{openai2025gpt52} to evaluate each trajectory along two criteria: (1) Information Recognition, whether the agent accurately recognizes what information has already been obtained; and (2) Gap Identification, whether the agent correctly identifies what specific information remains missing. The full evaluation prompt is provided in Appendix~\ref{appendix:ssa_eval_prompt}. Based on these criteria, we categorize each trajectory into one of three levels:

\begin{itemize}
    \item \textbf{Correct SSA}: The agent satisfies both criteria and provides the correct final answer.
    \item \textbf{Hallucinated SSA}: The agent satisfies both criteria but provides an incorrect answer.
    \item \textbf{Incorrect SSA}: The agent fails to satisfy at least one criterion, indicating a lack of awareness regarding its search state.
\end{itemize}

The results in Figure~\ref{figure:ka_barplot} show that SARL outperforms all baselines on search-state awareness. SARL achieves a Correct SSA rate of 42\%, compared to 24\% for Search-R1 and 17\% for both AutoRefine and ReSeek. The gap is similarly reflected in Incorrect SSA, where SARL fails on 51\% of trajectories versus 65--76\% for baselines, indicating that SARL reduces cases where the agent searches without recognizing what it has obtained or what remains missing. Notably, AutoRefine and ReSeek (both of which incorporate explicit refinement or judgment tokens during training) exhibit higher Incorrect SSA than Search-R1, suggesting that auxiliary reflection tokens alone do not internalize search-state awareness.

We note that the Correct SSA metric is stringent, requiring the agent to simultaneously recognize obtained information, identify missing gaps, and produce the correct final answer; even partial failures count as Incorrect SSA. The substantial gap between SARL and baselines under this strict criterion empirically supports our central premise that SARL's training objective produces meaningfully better search-state awareness, while leaving room for further improvement.

\subsection{Hyperparameter Sensitivity}
\label{appendix:hyperparam}

We provide the full sensitivity results for the assessment penalty coefficient $\alpha$ and reward coefficient $\lambda$ referenced in Section~\ref{subsec:reward_ablation}. Table~\ref{tab:hyperparam_sensitivity} reports the best and final validation scores under five configurations, varying one coefficient while holding the other fixed. Performance remains stable across $\alpha \in \{0.2, 0.5, 0.8\}$ when $\lambda=0.5$ is fixed, while $\lambda=0.2$ with $\alpha=0.8$ collapses sharply between best and final scores (55.5 to 42.6), indicating that under-weighting the assessment reward leads to training instability.

\begin{table}[h]
\centering
\small
\renewcommand{\arraystretch}{1.1}
\caption{Sensitivity of SARL to the assessment penalty ($\alpha$) and reward ($\lambda$) coefficients. Best and Final denote the highest and end-of-training validation scores, respectively. Default configuration ($\alpha=0.8, \lambda=0.5$) yields the strongest result.}
\label{tab:hyperparam_sensitivity}
\begin{tabular}{c c c c}
\toprule
$\alpha$ & $\lambda$ & \textbf{Best} & \textbf{Final} \\
\midrule
0.2 & 0.5 & 52.2 & 51.7 \\
0.5 & 0.5 & 55.0 & 54.8 \\
\rowcolor{gray!15}
0.8 & 0.5 & \textbf{56.3} & \textbf{55.5} \\
0.8 & 0.2 & 55.5 & 42.6 \\
0.8 & 0.8 & 54.5 & 54.5 \\
\bottomrule
\end{tabular}
\end{table}

\subsection{Generalization across Model Scale and Family}
\label{appendix:cross_model}

We provide per-step validation scores for the cross-model and cross-scale experiments referenced in Section~\ref{subsec:generalizability}. Table~\ref{tab:cross_model_llama} compares SARL against GRPO when applied to Llama-3.1-8B-Instruct: GRPO declines from an average of 52.3 to 44.4 over training, while SARL improves from 51.2 to 54.3, outperforming GRPO on every benchmark by step 300. Table~\ref{tab:cross_model_14b} compares SARL with the same Qwen2.5 backbone at 7B and 14B parameters: the 14B variant outperforms 7B across all benchmarks and training steps, with a particularly large gap on ASearcher.

\begin{table}[h]
\centering
\small
\caption{Generalization of SARL across model family (left) and scale (right).}
\label{tab:cross_model}
\begin{minipage}{0.48\columnwidth}
\centering
\setlength{\tabcolsep}{2pt}
\begin{tabular}{l l c c c c}
\toprule
\textbf{Step} & \textbf{Method} & \textbf{NQ} & \textbf{HQA} & \textbf{ASearcher} & \textbf{Avg.} \\
\midrule
\multirow{2}{*}{100} & GRPO & 48.4 & 58.2 & 50.4 & 52.3 \\
& SARL & 44.8 & 58.6 & 50.2 & 51.2 \\
\midrule
\multirow{2}{*}{200} & GRPO & 49.2 & 58.2 & 46.6 & 51.3 \\
& SARL & 48.2 & 56.6 & 49.6 & 51.5 \\
\midrule
\multirow{2}{*}{300} & GRPO & 40.8 & 54.6 & 37.8 & 44.4 \\
& \cellcolor{gray!15} SARL & \cellcolor{gray!15} \textbf{50.6} & \cellcolor{gray!15} \textbf{62.0} & \cellcolor{gray!15} \textbf{50.2} & \cellcolor{gray!15} \textbf{54.3} \\
\bottomrule
\end{tabular}
\subcaption{Llama-3.1-8B-Instruct}
\label{tab:cross_model_llama}
\end{minipage}
\hfill
\begin{minipage}{0.48\columnwidth}
\centering
\setlength{\tabcolsep}{2pt}
\begin{tabular}{l l c c c c}
\toprule
\textbf{Step} & \textbf{Scale} & \textbf{NQ} & \textbf{HQA} & \textbf{ASearcher} & \textbf{Avg.} \\
\midrule
\multirow{2}{*}{100} & 7B  & 44.0 & 52.2 & 41.0 & 45.7 \\
& 14B & 46.4 & 59.0 & 51.0 & 52.1 \\
\midrule
\multirow{2}{*}{200} & 7B  & 49.2 & 56.0 & 48.6 & 51.3 \\
& 14B & 50.8 & 62.0 & 55.8 & 56.2 \\
\midrule
\multirow{2}{*}{300} & 7B  & 49.4 & 59.8 & 53.0 & 54.1 \\
& \cellcolor{gray!15} 14B & \cellcolor{gray!15} \textbf{50.4} & \cellcolor{gray!15} \textbf{64.0} & \cellcolor{gray!15} \textbf{58.6} & \cellcolor{gray!15} \textbf{57.7} \\
\bottomrule
\end{tabular}
\subcaption{Qwen2.5 7B/14B}
\label{tab:cross_model_14b}
\end{minipage}
\end{table}

\subsection{Prompt Details}
\label{app:search_integrity_prompt}

\textbf{Rollout Generation.} We use the following prompt for rollout generation.

\begin{lstlisting}[
  basicstyle=\ttfamily\footnotesize,
  breaklines=true,
  breakatwhitespace=true,
  columns=fullflexible
]
Answer the given question. You must conduct reasoning inside <think> and </think> first every time you get new information. After reasoning, if you find you lack some knowledge, you can call a search engine by <search> query </search> and it will return the top searched results between <information> and </information>. You can search as many times as you want. If you find no further external knowledge needed, you can directly provide the answer inside <answer> and </answer>, without detailed illustrations. For example, <answer> Beijing </answer>. Question:
\end{lstlisting}


\paragraph{Self-Assessment Prompt.}
\label{appendix:assessment_prompt}
We use the following prompt to encourage the model to make self-assessments of its own search process. The prompt assesses whether all information required to answer a question was systematically sought before committing to a conclusion.
\begin{lstlisting}[
  basicstyle=\ttfamily\footnotesize,
  breaklines=true,
  breakatwhitespace=true,
  columns=fullflexible
]
The following is a search process that uses <think> to reason and <search> to collect evidence.
Your task is to judge the integrity of the search process itself, NOT whether the final answer is correct.

A search process has integrity only if it demonstrates that all information required to answer the question was systematically and transparently sought before committing to a conclusion.

Evaluate the process using the criteria below.

--------------------------------
Evaluation Criteria
--------------------------------

- An essential constraint is a minimal factual dependency that must be resolved in order to compute the answer.
- Direct questions have a single essential constraint; multi-step questions require multiple essential constraints.
- Each essential constraint must be constructed directly from the question, not from the reasoning or search behavior.
- The reasoning behavior must explicitly attempt to resolve each essential constraint.
- The reasoning behavior must not commit to a candidate answer before all essential constraints have been addressed.
- Explicit acknowledgment of missing information followed by an answer counts as failure.
- Any missing, skipped, implicitly assumed, or prematurely resolved constraint results in failure.

--------------------------------
Output Requirements
--------------------------------

Your response MUST follow this exact order:

1. Essential Constraint(s):
   - List each essential constraint implied by the question as short, minimal, concrete items.
   - Do NOT include any additional constraint that is not directly derived from the question.
   - The provided search process must not affect the construction of the essential constraint(s).

2. Analysis:
   - Analyze the integrity of the search process based on the criteria defined above.
   - Analyze whether the search process attempted to resolve each essential constraint before committing to an answer.

3. Judgment:
   - Output exactly one word: COMPLETE or INCOMPLETE

--------------------------------
Important Restrictions
--------------------------------

- Focus solely on the integrity of the search process.
- DO NOT perform any searches yourself.
- DO NOT answer the question.
- DO NOT assume facts that are not explicitly supported by the search process.
- If there is reasonable doubt that all essential constraints were properly addressed,
  you MUST choose INCOMPLETE.

--------------------------------
Search Process:
--------------------------------

{search_process}
\end{lstlisting}

\paragraph{Revision Instruction.}
\label{appendix:revision_prompt}
When the self-assessment yields an \texttt{INCOMPLETE} judgment, the assessment is concatenated with the following instruction and injected as a user message into the original chat history, prompting the model to revise its search process.
\begin{lstlisting}[
  basicstyle=\ttfamily\footnotesize,
  breaklines=true,
  breakatwhitespace=true,
  columns=fullflexible
]
Feedback:
{full_feedback}

Now, revise your search process based on the feedback above. You must first conduct reasoning inside <think> and </think>, discussing the validity of the given feedback. After reasoning, if you find you lack some knowledge, you can call a search engine by <search> query </search> and it will return the top searched results between <information> and </information>. You can search as many times as your want. If you find no further external knowledge needed, you can directly provide the answer inside <answer> and </answer>, without detailed illustrations. For example, <answer> Beijing </answer>. Reasoning:
\end{lstlisting}

\paragraph{LLM-as-a-Judge Prompt.}
\label{appendix:judge_prompt}

For evaluation, we use GPT-4.1 (\texttt{gpt-4.1-2025-04-14}) as the judge model. We adopt the judge prompt from BrowseComp-Plus, shown below.

\begin{lstlisting}[
  basicstyle=\ttfamily\footnotesize,
  breaklines=true,
  breakatwhitespace=true,
  columns=fullflexible
]
Judge whether the following [response] to [question] is correct or not based on the precise and unambiguous [correct_answer] below.

[question]: {question}

[response]: {pred_answer}

[correct_answer]: {gt_answer}

Your judgement must be in the format and criteria specified below:

extracted_final_answer: The final exact answer extracted from the [response].

[correct_answer]: Repeat the [correct_answer] given above.

reasoning: Explain why the extracted_final_answer is correct or incorrect based on [correct_answer], in the context of this [question]. You should judge whether the extracted_final_answer is semantically equivalent to [correct_answer], allowing the extracted_final_answer to be string variations of [correct_answer]. You should also allow the extracted_final_answer to be more precise or verbose than [correct_answer], as long as its additional details are correct. Do not comment on any background to the problem, do not attempt to solve the problem, do not argue for any answer different than [correct_answer], focus only on whether the answers are semantically equivalent.

correct: Answer 'yes' if extracted_final_answer matches the [correct_answer] given above, or is within a small margin of error for numerical problems. Answer 'no' otherwise, i.e. if there is any inconsistency, ambiguity, non-equivalency, or if the extracted answer is incorrect.

confidence: The extracted confidence score between 0% and 100% from [response]. Put 100 if there is no confidence score available.
\end{lstlisting}

\paragraph{Search-State Awareness Evaluation Prompt.}
\label{appendix:ssa_eval_prompt}
We use the following prompt with GPT-5.2 to evaluate the search-state awareness of agent trajectories along the criteria defined in Appendix~\ref{subsec:preliminary_results}.

\begin{lstlisting}[
  basicstyle=\ttfamily\footnotesize,
  breaklines=true,
  breakatwhitespace=true,
  columns=fullflexible
]
You are an expert evaluator assessing the search-state awareness of an AI search agent.

You will be given:
1. A question the agent must answer
2. The agent's full search trajectory (reasoning, search queries, and search results)
3. The ground truth answer

Your task is to evaluate the agent's reasoning trajectory on two criteria:

## Criteria

### 1. Information Recognition
Has the agent accurately recognized what information has already been obtained from search results?
- Score 1 if: The agent correctly acknowledges and references information it has retrieved. It does not ignore or contradict information already obtained.
- Score 0 if: The agent fails to recognize information it has already obtained, ignores relevant retrieved information, or misrepresents what it has found.

### 2. Gap Identification
Has the agent correctly identified what specific information remains missing?
- Score 1 if: The agent correctly identifies what additional information is needed to answer the question and searches for it, OR correctly determines that sufficient information has been gathered (no gap remains).
- Score 0 if: The agent fails to identify missing information, hallucinates that it has information it does not, or prematurely concludes without recognizing gaps in its knowledge.

## Important Notes
- Focus on the agent's REASONING process (inside <think> tags), not just the final answer.
- Evaluate whether the agent demonstrates awareness of its search state throughout the trajectory.
- If search results already contain sufficient information, recognizing that no gap remains counts as correct gap identification.
- Do NOT let the correctness of the final answer influence your assessment of the two criteria. A wrong answer can still show good search-state awareness, and a correct answer can still show poor awareness.

## Output Format
You MUST respond in the following exact JSON format:
{
  "information_recognition": {
    "score": 0 or 1,
    "reasoning": "Brief explanation"
  },
  "gap_identification": {
    "score": 0 or 1,
    "reasoning": "Brief explanation"
  }
}

---

## Question
{question}

## Ground Truth Answer
{answer}

## Agent's Search Trajectory
{trajectory}
\end{lstlisting}

\subsection{Broader Impacts}
\label{appendix:broader_impacts}

We discuss broader impacts from two perspectives: the implications of search-augmented agents trained with internalized self-assessment, and considerations specific to the SARL training framework.

\paragraph{Search-augmented agents.}
Search agents that can identify and recover from incomplete reasoning have the potential to substantially improve the reliability of retrieval-augmented systems. By learning to recognize when retrieved evidence is insufficient, such agents reduce the rate of confident but unsupported answers, which is a major failure mode of current LLMs deployed in information-seeking applications. Positive applications include research assistance, education, healthcare information access, and decision support, where the cost of fabricated or unverified outputs is high.

However, more capable search agents also raise familiar concerns. Agents that produce plausible, well-formatted answers may amplify misinformation when retrieved evidence is itself biased, outdated, or inaccurate, a risk that grows when users implicitly trust agent outputs without verifying sources. Malicious actors could exploit such systems for targeted information extraction or for synthesizing convincing disinformation at scale. We view rigorous evaluation, transparent source attribution, and bounded retrieval corpora as practical mitigations during deployment, none of which the underlying training method alone can guarantee.

\paragraph{The SARL framework.}
SARL is a training method rather than a deployed system, and is built entirely on publicly available base models, retrievers, and benchmarks. It does not introduce capabilities that meaningfully expand the misuse surface of the underlying backbones; the agents it produces inherit the risk profile of the base model and the retrieval corpus. A specific consideration when deploying capable search agents is that improved coherence and reliability of outputs can encourage users to trust them more uncritically. This reinforces the importance of pairing capability gains with explicit uncertainty signaling and source verification at the application layer.

\end{document}